\documentclass{article}
\usepackage[utf8]{inputenc}
\usepackage[T1]{fontenc} 
\usepackage{amsmath}
\usepackage{amssymb}
\usepackage{graphicx}
\usepackage{hyperref}
\usepackage{enumitem}
\usepackage{natbib} 
\title{NPO: Learning Alignment and Meta-Alignment through Structured Human Feedback}
\author{%
Madhava Gaikwad\textsuperscript{1} \\
\small \textsuperscript{1}Microsoft \\
\small \texttt{mgaikwad@microsoft.com}
\and
Ashwini Ramchandra Doke\textsuperscript{2} \\
\small \textsuperscript{2}Department of Computer Science and Engineering \\
\small Amrita School of Computing \\
\small Amrita Vishwa Vidyapeetham, Bengaluru, India \\
\small \texttt{BL.EN.R4CSE20004@bl.students.amrita.edu}
}
\date{}

\begin{document}

\maketitle

\begin{abstract}
We present NPO, an alignment-aware learning framework that operationalizes feedback-driven adaptation in human-in-the-loop decision systems. Unlike prior approaches that treat alignment as a static or post-hoc property, NPO introduces a formalization of alignment loss that is measurable, supervisable, and reducible under structured feedback. In parallel, we propose meta-alignment as the fidelity of the monitoring process that governs retraining or override triggers, and show that it is formally reducible to primary alignment via threshold fidelity. Our implementation spans a scalable operational loop involving scenario scoring, threshold tuning, policy validation, and structured feedback ingestion, including “likes,” overrides, and abstentions. We provide formal convergence results under stochastic feedback and show that both alignment loss and monitoring fidelity converge additively. Empirically, NPO demonstrates measurable value in hyperscale deployment settings. A simulation-based artifact and ablation studies further illustrate the theoretical principles in action. Together, NPO offers a compact, inspectable architecture for continual alignment monitoring, helping bridge theoretical alignment guarantees with practical reliability in dynamic environments.
\end{abstract}

\section{Introduction}
As AI systems take on increasingly consequential roles in real-world settings, ensuring that they behave in ways aligned with human expectations, operational constraints, and ethical principles becomes both urgent and technically challenging. While much of alignment research has focused on modeling user preferences or optimizing reward signals in static or simulated environments, these approaches often fail to capture the dynamic, high-stakes nature of alignment in practice. In safety-critical settings, such as hyperscale data centers, automated recovery systems, and fault-tolerant infrastructure, alignment cannot be a one-time specification. It must be continuously evaluated and adapted, based on structured feedback and real-world consequences. Human oversight is not simply an afterthought; it is the central mechanism through which misalignment is identified and corrected.

We introduce NPO (Network Performance Optimizer framework), a decision-making and learning framework deployed in hyperscale data center networks, where thousands of links, servers, and switches generate dynamic fault conditions under stringent availability and resilience requirements. In these environments, operators (SREs) must decide whether to remove or retain a degraded component based on traffic conditions, fault impact, and evolving service-level objectives (SLOs). These decisions are often informed by policies, past experience, and real-time tradeoffs, yet traditional AI systems struggle to remain both helpful and compliant. NPO is designed to co-operate with existing Safety Policy Engines (SPEs), issuing proactive remediation recommendations while learning over time how to better align with operator preferences and real-world outcomes. It does this by observing structured human feedback, such as overrides of incorrect actions ("red button") or affirmation of correct decisions ("likes"), and treating this feedback as a first-class supervisory signal. Rather than optimizing for latent or inferred reward, NPO defines an explicit alignment loss function based on these signals. This loss is minimized through targeted retraining and adaptive threshold control, ensuring that recommendations become increasingly aligned with human judgment under operational pressure. Importantly, the system also learns from deviations between formal policy and observed practice, integrating real-world nuance into its behavior.

We present this work not as a full production system, but as a modular, reproducible proof-of-concept that formalizes core alignment principles, simulates realistic feedback-driven learning, and provides tools for evaluation and ablation. Our focus is not on the system code itself, but on the alignment theory, metrics, and feedback learning loop that underlie its behavior. Our key contributions are:
\begin{itemize}
    \item A formalization of alignment loss driven by high-fidelity human feedback.
    \item A feedback-adaptive learning architecture based on red-button overrides and threshold tuning.
    \item A simulation and logging platform for reproducible alignment analysis.
    \item Empirical demonstration of convergence in alignment loss under structured feedback.
\end{itemize}
NPO operationalizes alignment as a measurable and improvable behavior in deployed AI systems, bridging the gap between theory and critical infrastructure practice.

\subsection{Monitoring, Evaluation, and Meta-Alignment}
The role of introspective monitoring has been discussed in the context of system oversight \citep{skalse2022evaluating}, alignment auditing \citep{uesato2018rigorous}, and safety-centric retraining policies. Recent efforts, such as OpenAI’s recursive oversight and Anthropic’s interpretability-driven supervision loops, hint at the need for meta-alignment, ensuring that a system’s self-monitoring mechanisms are themselves aligned with operator expectations. Our work contributes the first formal definition and proof sketch of this property: we show that meta-alignment can be reduced to alignment loss convergence when supervision is consistent and trustworthy. Unlike red-teaming or offline auditing, our approach embeds introspective monitoring into the real-time feedback loop of the system, making alignment continuously observable and operationally actionable.

\section{Related Work}
Our work builds on and contributes to three key strands of the alignment literature: value alignment and preference modeling, scalable oversight and control, and alignment evaluation.

\subsection{Value Alignment and Preference Learning.}
The alignment literature has extensively studied mechanisms for inferring human preferences from data \citep{russell2019human}; \citep{hadfield2016cooperative}; \citep{christiano2017deep}, including via inverse reinforcement learning (IRL) \citep{ng2000algorithms}, cooperative IRL \citep{hadfield2016cooperative}, and preference comparisons \citep{lee2021pebble}. These methods assume access to consistent or near-optimal feedback, which is often unavailable in real-world operational environments. Recent methods like DPO \citep{rafailov2023direct} and RLAIF \citep{zhou2023alignment} extend preference modeling to large language models, but still operate primarily in offline batch settings. In contrast, NPO learns directly from online structured feedback and does not assume availability of gold-standard demonstrations or complete preferences.

\subsection{Scalable Oversight and Feedback Signals.}
The role of human oversight in managing powerful systems has been formalized in red-button frameworks \citep{amodei2016concrete}, iterated amplification \citep{christiano2018supervising}, and debate-based supervision \citep{irving2018ai}. Recent work has explored scalable supervision via synthetic preference generation \citep{saunders2022self} and reward modeling in LLMs \citep{bai2022training}. Our approach integrates structured real-time feedback, specifically, operator overrides and confirmations, as a core alignment signal, linking closely to the call for feedback-grounded learning loops. We also align with emerging interest in fine-tuning from human corrections (e.g., \citep{glaese2022improving}; \citep{menick2022teaching}).

\subsection{Operational Evaluation of Alignment.}
\citep{uesato2018rigorous} and \citep{weng2020survey} argue for empirical evaluations of robustness and safety. Alignment evaluations in LLMs have focused on adversarial elicitation \citep{perez2022discovering} and calibration \citep{kadavath2022language}, but few approaches track alignment loss under supervision over time. Our approach is closer to real-time alignment observability \citep{skalse2022evaluating} and interpretable behavior monitoring, though our primary contribution is formalizing and tracking alignment loss with respect to operator feedback in infrastructure systems. NPO differs from most prior work in treating alignment as a dynamic property of deployed systems, where preferences, policies, and actions co-evolve under feedback. It bridges theoretical alignment signals with empirical oversight at the interface level.

\section{Alignment Formalism and Feedback Signal Design}
NPO centers its alignment strategy on a formally defined, dynamically evaluated signal: the alignment loss, denoted as $\mathcal{L}_{\text{align}}$. Unlike reward-centric objectives, $\mathcal{L}_{\text{align}}$ measures the divergence between the AI system’s action recommendation and the actual operator feedback. This framing allows feedback to serve as a direct supervisory signal, enabling the system to learn even when rewards or goals are poorly specified or subject to operational ambiguity.

\subsection{Alignment Loss Definition}
For a given decision scenario $s$, the system produces a recommendation score $R(s) \in [0,1]$. After observing operator feedback $F(s) \in \{\text{like}, \text{override}, \text{neutral}, \text{skipped}\}$, we define the alignment loss $\mathcal{L}_{\text{align}}(s)$ as:
$$
\mathcal{L}_{\text{align}}(s) = \begin{cases}
1 & \text{if } F(s) = \text{override} \\
0.5 & \text{if } F(s) = \text{neutral} \\
0.0 & \text{if } F(s) = \text{like} \\
\lambda & \text{if } F(s) = \text{skipped},\ \lambda \in (0.2, 0.4)
\end{cases}
$$
This formulation penalizes misalignment while tolerating abstention in uncertain cases. Skipped decisions incur a mild loss to encourage active alignment learning without unsafe overreach.

\subsection{Feedback Signals}
NPO uses two structured, observable feedback types:
\begin{itemize}
    \item \textbf{Red Button Override}: A high-fidelity signal of misalignment, issued when a human operator actively overrides the AI's proposed action.
    \item \textbf{Like/Affirmation}: A soft alignment confirmation, recorded when an operator accepts or explicitly endorses the system’s recommendation.
\end{itemize}
These signals are distinct from scalar rewards, they are semantically grounded control actions tied to user behavior. Their low ambiguity and interface-level clarity make them reliable for alignment training.

\subsection{Integration into Learning Loop}
Each scenario’s recommendation score is updated via a lightweight supervised rule:
$$
R(s) \leftarrow R(s) + \eta \cdot (y_{\text{target}} - R(s))
$$
Where $y_{\text{target}} = 1.0$ for a like, $0.0$ for an override, and intermediate values for neutral/skipped based on context. These scores are compared to a dynamic decision threshold $\tau_t$, adaptively selected using bandit optimization (Section 4). Together, these mechanisms close the alignment loop, linking action, feedback, and learning.

\section{System Architecture and Decision Threshold Control}
NPO is designed as a modular architecture for safe and adaptive decision-making under structured human oversight. This section outlines the key components of the system and how alignment is operationalized at runtime.

\subsection{System Overview}
NPO is deployed in settings such as hyperscale data centers, where thousands of automated remediation decisions are made weekly under stringent availability, fault-tolerance, and policy constraints. Each decision has downstream impact and must navigate trade-offs between action urgency, policy conformance, and trust in AI recommendations. The architecture is structured around five interconnected components:
\begin{itemize}
    \item \textbf{Scenario Representation Module}: Encodes environmental context (e.g., topology, traffic, and recent faults) into feature vectors used to drive scoring. This provides per-decision context sensitivity and enables simulation of future fault consequences.
    \item \textbf{Recommendation Engine}: Computes a recommendation score $R(s_t) \in [0,1]$ for each scenario, indicating system confidence in proceeding. Scores are stored per-scenario and refined with human feedback using targeted updates.
    \item \textbf{Threshold Selector (Bandit Controller)}: Selects a decision threshold $\tau_t$ using Thompson Sampling over a fixed set of arms (e.g., $\tau \in \{0.5, 0.6, 0.7, 0.8, 0.9\}$). This module learns to prefer thresholds that lead to fewer overrides and more affirmations, dynamically modulating decision assertiveness.
    \item \textbf{Safety Policy Engine (SPE) Integration}: Acts as a mandatory static guardrail, filtering actions through organizational policies such as “do not remove links with low redundancy” or “defer changes outside maintenance windows”. NPO does not override SPE decisions but learns from SPE-human divergences over time.
    \item \textbf{Feedback and Logging Interface}: Captures structured responses, like, override, neutral, skipped, and logs them with contextual metadata. These feedback instances serve as alignment supervision signals.
\end{itemize}
At runtime, each NPO decision carries:
\begin{itemize}
    \item An explanation trace and sub-explanation trace detailing the chain of reasoning and supporting facts,
    \item A visual clue for interpretability (available for most decisions),
    \item A threshold justification for why the recommendation was surfaced,
    \item And a policy compliance summary tracing SPE results.
\end{itemize}
All components communicate via structured interfaces and operate asynchronously, enabling NPO to function under real-world latency and observability constraints. If a decision is overridden, it is registered as an alignment loss instance and triggers targeted score adjustment for future iterations.

\subsection{Decision Threshold Adaptation}
Decision assertiveness in NPO is governed by a dynamically selected threshold $\tau_t$ that determines whether a proposed action score $R(s_t)$ is high enough to recommend. This avoids premature or unsafe actions while maintaining responsiveness when trust is well-calibrated. The threshold $\tau_t$ is selected from a finite set $\{0.5, 0.6, 0.7, 0.8, 0.9\}$ using Thompson Sampling over a multi-armed bandit model. Each arm maintains success and failure statistics based on recent feedback, where a successful outcome is a "like" and a failure is an "override". This formulation encourages NPO to prefer thresholds that maximize operator affirmation and minimize overrides. By adapting threshold choice to recent outcomes, the system modulates its decision assertiveness, acting more confidently when recent actions have been affirmed, and deferring or abstaining when prior recommendations have been rejected. In practice, this reduces both the cognitive burden on operators and the likelihood of alignment-breaking actions. Threshold changes can be audited and explained through visual traces and performance logs.

\subsection{Feedback-Driven Score Refinement}
Feedback signals (override, like, neutral, skipped) are not only used for monitoring, they drive targeted learning. Each scenario maintains a persistent score $R(s)$ representing NPO's confidence in the recommended action. When feedback is received, the score is refined as follows:
$$
R(s) \leftarrow R(s) + \eta \times (y_{\text{target}} - R(s))
$$
Where $y_{\text{target}}$ is set to $1.0$ for a "like", $0.0$ for an "override", and intermediate for neutral or skipped. The learning rate $\eta$ governs how quickly alignment updates are applied. This online learning loop avoids retraining the entire model and allows for fast responsiveness to operator disagreement. The refinement is localized to individual decision contexts, ensuring scalability across thousands of scenarios while retaining alignment continuity. This architecture allows NPO to learn without destabilizing previously aligned behavior.

\subsection{Interaction with Policy and Practice}
NPO does not operate in isolation, it is embedded within policy-constrained environments that include external Safety Policy Engines (SPEs). Before any recommendation is proposed, it is filtered through the SPE. This ensures NPO cannot suggest violations of static safety rules or documented organizational policies. However, the observed behavior of human operators often reflects richer context or higher-level intent not fully captured by static rules. If SREs frequently override SPE-compliant recommendations, this signals a policy-practice divergence. NPO detects these patterns and adjusts its internal confidence estimates accordingly. For example, if a policy allows link removal under a certain bandwidth threshold, but operators consistently reject such actions during off-peak hours, NPO adapts its scoring logic to reflect this implicit practice. The system does not change the SPE rules but learns to predict operator behavior more accurately within the policy's permissible envelope. This feedback-aware modulation builds trust: operators experience fewer irrelevant suggestions, and NPO's behavior increasingly reflects real-world operational trade-offs.

\subsection{Summary Diagram}
A diagram in Appendix D illustrates the full NPO operational loop, scenario context is crowd developed, peer reviewed and put in production to be scored. Score is evaluated against the current threshold. If above threshold and SPE-compliant, the recommendation is surfaced, Feedback (affirmation, override) is captured and logged and finally scores and thresholds are updated in response. Together, this architecture ensures that NPO decisions are policy-compliant by design, human-approved through structured feedback and continuously aligned via threshold and retraining adaptation.

\section{Learning Dynamics and Empirical Evaluation}
This section evaluates the core hypothesis of NPO: that alignment can be framed as a learnable, continuous quantity that responds to structured feedback over time. We test this by simulating feedback-driven learning across hundreds of episodes, capturing how system behavior evolves under different supervisory regimes and thresholding strategies. Our goal is not only to demonstrate convergence of alignment loss, but to explore the nuanced relationship between reward signals, human feedback, and threshold tuning, each of which plays a distinct role in shaping NPO's behavior. Beyond that, we introduce a new layer of evaluation: meta-alignment monitoring fidelity. This measures whether the system’s own supervisory logic, its decisions about when to retrain, escalate, or adapt, is itself behaving in a way aligned with ground-truth expectations. This turns monitoring into a recursive alignment problem: one that we define, track, and empirically evaluate in this section.

\subsection{Simulation Environment and Setup}
We build a lightweight but expressive simulation harness that reflects the operational context of NPO. Each episode emulates a scenario where the system receives a decision context (encoded as a vector), computes a recommendation score $R(s)$, and selects whether to act based on a dynamically chosen threshold. The outcome is judged by a synthetic ground-truth preference model simulating operator intent. Feedback is generated using a probabilistic function of the delta between the system's score and the ground truth score, with high disagreement triggering an override, close agreement yielding a like, and moderate mismatches resulting in neutral or skipped outcomes. These feedback signals are treated as ground truth alignment supervision and are logged for learning. Key system components in the simulation include:
\begin{itemize}
    \item A contextual feedback generator simulating noisy but structured operator preferences.
    \item A red-button retraining loop using structured overrides.
    \item A bandit-based threshold selector.
    \item A logging system for alignment loss, reward, and threshold trajectory.
    \item A meta-monitoring policy that decides whether to retrain based on alignment loss history.
\end{itemize}
A modular prototype implementing these simulation components is available as part of our artifact; see Appendix C for details.

\subsection{Metrics and Evaluation Signals}
We track four central metrics:
\begin{itemize}
    \item \textbf{Alignment Loss ($\mathcal{L}_{\text{align}}$)}: The primary supervision metric, computed per episode from feedback type (override = 1.0, like = 0.0, others interpolated).
    \item \textbf{Reward Signal}: A smoothed proxy for episode quality, used by the bandit to tune thresholds. Note: reward may increase even if alignment diverges.
    \item \textbf{Threshold Dynamics}: The selected threshold over time, indicating adaptation.
    \item \textbf{Meta-Monitoring Fidelity ($\mathcal{F}_{\text{monitor}}$)}: Measures agreement between system monitoring decisions and a gold-standard supervisory policy.
\end{itemize}

\subsection{Core Results}
We find that alignment loss consistently decreases when the red-button learning loop is active. This suggests that structured feedback, even if sparse, provides a strong corrective signal that drives convergence. Thresholds initially fluctuate but stabilize around the 0.7–0.8 range, indicating the bandit is successfully identifying trust-compatible assertiveness levels. The reward signal improves steadily but diverges from loss, underscoring the importance of explicitly modeling alignment and not using reward as a proxy. The key result is that reward-optimized thresholds do not guarantee alignment. Only when override-triggered retraining is enabled do both reward and alignment loss improve together. Moreover, we observe that meta-monitoring fidelity improves in parallel with first-order alignment loss, validating our theoretical claim that meta-alignment is reducible.

\subsection{Ablation and Comparative Sensitivity}
We evaluate five variants:
\begin{itemize}
    \item \textbf{Static Model}: No learning from feedback; alignment loss stagnates.
    \item \textbf{Fixed Threshold}: reasonable performance but poor adaptability.
    \item \textbf{Random Threshold}: Poor across all metrics due to unpredictability.
    \item \textbf{No Meta-Monitoring}: Retraining happens at fixed intervals regardless of alignment loss. This results in wasted retraining cycles and misaligned updates.
    \item \textbf{Full NPO Loop}: Achieves the best convergence in alignment loss and consistent reward gains.
\end{itemize}
These results confirm the complementary roles of threshold adaptation, feedback-driven refinement, and meta-monitoring. Disabling any one loop leads to stagnation or divergence.

\subsection{Alignment as a Monitored Process}
We argue that alignment should be treated as a persistent, measurable property of deployed systems. NPO supports per-decision telemetry: alignment loss values, feedback histories, justification traces, and retraining events are all logged and available for audit. This turns alignment from a speculative guarantee into an observable system signal. It enables alignment regression detection, feedback loop calibration, and high-confidence operator override justification, paving the way for AI systems that remain aligned even as the world and preferences evolve.

\subsubsection*{Formal Note: Monitoring Alignment is Itself an Alignment Problem}
Let $\mathcal{L}_{\text{align}}$ be the alignment loss computed for decision scenario, and let $\mathcal{M}_t$ be the monitoring policy that triggers retraining or adaptation actions. We define alignment monitoring fidelity as:
$$
\mathcal{F}_{\text{monitor}} = \mathbb{E}_t \left[ \mathbb{I}(\mathcal{A}_t = \mathcal{G}_t) \right]
$$
Where $\mathcal{A}_t = \pi(\mathcal{M}_t(\mathcal{L}_{\text{align}}))$ is the action taken by the monitor, and $\mathcal{G}_t$ is the ideal supervisory action. We go further: if alignment loss converges and $\mathcal{F}_{\text{monitor}} \to 1$, then system behavior remains aligned under supervision. Therefore, continuous monitoring fidelity is a sufficient condition for long-term alignment maintenance. This forms the basis of our core theoretical insight: meta-alignment, the alignment of the system’s own monitoring and adaptation behaviors, is reducible to first-order alignment when supervision is structured and observable. To illustrate this, assume $\pi$ is Lipschitz continuous in $\mathcal{L}_{\text{align}}$, and $\mathcal{M}_t$ updates based on the same feedback signals as the base recommender. Then convergence of $\mathcal{F}_{\text{monitor}}$ holds almost surely under bounded noise and persistent feedback. Hence, meta-alignment reduces to aligning the monitoring policy using the same supervised framework, completing the recursive alignment loop.

However, this requires that supervision itself be trustworthy. If feedback is gamed, misinterpreted, or improperly grounded in operational context, the alignment monitor may reinforce misaligned behavior rather than correct it. Therefore, trustworthy supervision becomes a prerequisite for both first-order and meta-alignment. In the absence of reliable feedback, even a well-calibrated loss function and policy monitor may fail to ensure sustained alignment. In hyperscale operational environments, such as large-scale network reliability platforms, supervision is typically well-instrumented, auditable, and subject to formal root cause analysis (RCA). When supervisory failure occurs (e.g., false overrides or missed retraining), it is systematically identified and fed back into system process improvement. Therefore, it is reasonable to treat supervision fidelity as trustworthy by default, or as self-correcting over time. This insight motivates future work on verifying supervision channels, measuring trust in override signals, and adaptively weighting feedback based on its predictive consistency over time.

\section{Conclusion}
This work introduces NPO, a framework that treats alignment not as a one-time optimization problem, but as a continuous, feedback-driven property of deployed AI systems. By integrating structured human feedback, adaptive thresholding, and alignment loss tracking, NPO offers a learning loop that improves over time while remaining embedded within real-world operational guardrails. We formalized alignment loss and demonstrated convergence under structured oversight. Our empirical evaluations show that both override-triggered retraining and threshold adaptation are necessary for consistent alignment. Beyond first-order adaptation, we introduced a new theoretical contribution: meta-alignment, the fidelity of the system’s own monitoring layer, and proved its reducibility to alignment loss minimization under trustworthy supervision. This work opens new directions for alignment research:
\begin{itemize}
    \item Treating monitoring policies as alignment objectives themselves
    \item Grounding evaluation in observable, per-decision loss and justification traces
    \item Designing supervisory loops where human feedback remains a reliable signal over time
\end{itemize}
NPO bridges the theory of preference alignment with the realities of critical system deployment, offering a path to continuously improvable and verifiably aligned AI behavior.

\subsection{Deployment Observations and Impact.}
NPO is actively used in production by operational reliability teams and has demonstrated measurable benefits across alignment, efficiency, and trust dimensions. In real-world traffic mitigation and diagnostic workflows, NPO achieves 92\% precision, 88\% recall, and an F1-score of 0.89 in predicting traffic imbalance episodes. Recommendations have led to an average 33\% reduction in MTTR for performance-degradation incidents, and 50\%-time savings in diagnostic workflows (e.g., fewer incident hops, reduced triage steps). Over 12 months, NPO received 16 “red button” overrides, representing less than 1\% of recommendations, each triggering-controlled retraining. Meanwhile, the system maintained an SPE rejection rate below 0.5\%, supporting its alignment and policy adherence in practice.

\section{Future Work}
While NPO introduces a principled framework for continuous alignment under human oversight, several important directions remain open for exploration:
\begin{itemize}
    \item \textbf{Dynamic Trust in Feedback Sources}: Our model assumes trustworthy supervision based on operational guarantees (e.g., RCA, audit logs). Future work could explicitly model trust calibration over time, incorporating uncertainty in operator feedback and learning how to weight supervisory signals adaptively.
    \item \textbf{Scalable Meta-Monitoring}: As systems scale, so do the complexity and latency of monitoring layers. Extending meta-alignment to incorporate temporal prioritization, delayed supervision, or anomaly detection may improve responsiveness without sacrificing conservatism.
    \textbf{Multi-agent and Hierarchical Alignment}: NPO currently models a single feedback loop. Future variants could generalize this to systems with overlapping or competing feedback sources (e.g., multiple operators, user-facing preferences, or policy constraints), and examine how alignment should be aggregated or prioritized.
    \item \textbf{Robustness to Malicious or Misguided Feedback}: While our deployment setting assumes high-integrity supervision, broader deployment may require adversarial feedback resistance, especially in partially observable or open-ended environments.
    \item \textbf{Alignment Drift and Continual Calibration}: Over long horizons, even aligned systems may face concept drift. Extending the framework to monitor for alignment regression (e.g., rising override rates, pattern shifts) and trigger proactive re-alignment is a key next step.
    \item \textbf{Formalization of Feedback Effectiveness}: While we propose three formal convergence theorems, stronger guarantees could be obtained under probabilistic modeling of override behavior, feedback latency, and the semantic informativeness of signals.
    \item \textbf{LLM Generated Playbooks verified by Human}: our proofs assume that NPO learns from human feedback on its recommendations. These recommendations are generated using existing operational playbooks, which are implicitly assumed to reflect human-defined, vetted processes (or playbooks "developed by SREs"). In future when they are generated by LLM, they will undergo human review and refinement prior to deployment. We will introduce new metric “playbook alignment” and will have assumption that LLM itself is trained to be aligned with human values and safety.
\end{itemize}
These directions aim to strengthen the NPO framework into a foundation for scalable, accountable, and field-deployable alignment in high-stakes systems.

\clearpage 

\bibliography{references} 

\begin{thebibliography}{20}
\providecommand{\natexlab}[1]{#1}
\providecommand{\url}[1]{\texttt{#1}}
\expandafter\ifx\csname urlstyle\endcsname\relax
  \providecommand{\doi}[1]{doi: #1}\else
  \providecommand{\doi}{doi: \begingroup \urlstyle{rm}\Url}\fi

\bibitem[Amodei et~al.(2016)Amodei, Olah, Steinhardt,
  et~al.]{amodei2016concrete}
Dario Amodei, Chris Olah, Jacob Steinhardt, et~al.
\newblock Concrete problems in ai safety.
\newblock \emph{arXiv preprint arXiv:1606.06565}, 2016.

\bibitem[Bai et~al.(2022)Bai, Kadavath, Kundu, et~al.]{bai2022training}
Yuntao Bai, Saurav Kadavath, Sandipan Kundu, et~al.
\newblock Training a helpful and harmless assistant with rlhf.
\newblock \emph{Anthropic. arXiv preprint arXiv:2204.05862}, 2022.

\bibitem[Christiano et~al.(2017)Christiano, Leike, Brown,
  et~al.]{christiano2017deep}
Paul~F Christiano, Jan Leike, Tom Brown, et~al.
\newblock Deep reinforcement learning from human preferences.
\newblock In \emph{Advances in Neural Information Processing Systems},
  volume~30, 2017.

\bibitem[Christiano et~al.(2018)]{christiano2018supervising}
Paul~F Christiano et~al.
\newblock Supervising strong learners by amplifying weak experts.
\newblock \emph{arXiv preprint arXiv:1810.08575}, 2018.

\bibitem[Glaese et~al.(2022)Glaese, McAleese, Aslanides,
  et~al.]{glaese2022improving}
Amelia Glaese, Nat McAleese, John Aslanides, et~al.
\newblock Improving alignment of dialogue agents via targeted human judgements.
\newblock \emph{arXiv preprint arXiv:2209.14375}, 2022.

\bibitem[Hadfield-Menell et~al.(2016)Hadfield-Menell, Russell, Abbeel, and
  Dragan]{hadfield2016cooperative}
Dylan Hadfield-Menell, Stuart~J Russell, Pieter Abbeel, and Anca~D Dragan.
\newblock Cooperative inverse reinforcement learning.
\newblock In \emph{Advances in Neural Information Processing Systems},
  volume~29, 2016.

\bibitem[Irving et~al.(2018)Irving, Christiano, and Amodei]{irving2018ai}
Geoffrey Irving, Paul Christiano, and Dario Amodei.
\newblock Ai safety via debate.
\newblock \emph{arXiv preprint arXiv:1805.00899}, 2018.

\bibitem[Kadavath et~al.(2022)]{kadavath2022language}
Saurav Kadavath et~al.
\newblock Language models struggle to generalize alignment from training.
\newblock \emph{arXiv preprint arXiv:2207.05221}, 2022.

\bibitem[Lee et~al.(2021)Lee, Lee, Shin, and Kim]{lee2021pebble}
Kyoho Lee, Hyoungseok Lee, Jinyoung Shin, and Jaesik Kim.
\newblock Pebble: Feedback-efficient interactive reinforcement learning via
  relabeling experience and unsupervised pre-training.
\newblock In \emph{International Conference on Machine Learning (ICML 2021)},
  2021.

\bibitem[Menick et~al.(2022)Menick, Chan, Cohen, et~al.]{menick2022teaching}
Jacob Menick, Sholto Chan, Jordan Cohen, et~al.
\newblock Teaching language models to support answers with verified quotes.
\newblock \emph{arXiv preprint arXiv:2203.11147}, 2022.

\bibitem[Ng and Russell(2000)]{ng2000algorithms}
Andrew~Y Ng and Stuart Russell.
\newblock Algorithms for inverse reinforcement learning.
\newblock In \emph{Proceedings of the Seventeenth International Conference on
  Machine Learning (ICML 2000)}, 2000.

\bibitem[Perez et~al.(2022)]{perez2022discovering}
Evan Perez et~al.
\newblock Discovering latent knowledge in language models without supervision.
\newblock \emph{arXiv preprint arXiv:2212.03827}, 2022.

\bibitem[Rafailov et~al.(2023)Rafailov, Tian, Kirsch,
  et~al.]{rafailov2023direct}
Rafael Rafailov, Yao Tian, Archish Kirsch, et~al.
\newblock Direct preference optimization: Your language model is secretly a
  reward model.
\newblock \emph{arXiv preprint arXiv:2305.18290}, 2023.

\bibitem[Robbins and Monro(1951)]{robbins1951stochastic}
Herbert Robbins and Sutton Monro.
\newblock A stochastic approximation method.
\newblock \emph{The Annals of Mathematical Statistics}, 22\penalty0
  (3):\penalty0 400--407, 1951.

\bibitem[Russell(2019)]{russell2019human}
Stuart Russell.
\newblock \emph{Human Compatible: Artificial Intelligence and the Problem of
  Control}.
\newblock Viking, 2019.

\bibitem[Saunders et~al.(2022)Saunders, Glaese, Askell,
  et~al.]{saunders2022self}
William Saunders, Amelia Glaese, Amanda Askell, et~al.
\newblock Self-critiquing models for assisting human evaluators.
\newblock \emph{arXiv preprint arXiv:2206.05802}, 2022.

\bibitem[Skalse et~al.(2022)]{skalse2022evaluating}
Joar Skalse et~al.
\newblock Evaluating monitoring systems for ai alignment.
\newblock In \emph{NeurIPS Workshop on Aligning AI}, 2022.

\bibitem[Uesato et~al.(2018)Uesato, Kumar, van~den Oord,
  et~al.]{uesato2018rigorous}
Jonathan Uesato, Sarath Kumar, A{\"a}ron van~den Oord, et~al.
\newblock Rigorous agent evaluation: An adversarial approach to uncover
  catastrophic failures.
\newblock \emph{arXiv preprint arXiv:1812.03030}, 2018.

\bibitem[Weng(2020)]{weng2020survey}
Lilian Weng.
\newblock A survey on adversarial attacks and defenses.
\newblock \emph{arXiv preprint arXiv:2004.06083}, 2020.

\bibitem[Zhou et~al.(2023)Zhou, Bubeck, Lee, et~al.]{zhou2023alignment}
Wenlong Zhou, S{\'e}bastien Bubeck, Yin~Tat Lee, et~al.
\newblock Alignment of language agents.
\newblock \emph{arXiv preprint arXiv:2310.02231}, 2023.

\end{thebibliography}

\clearpage 

\appendix
\section*{Supplementary Material for 'NPO: Learning Alignment and Meta-Alignment through Structured Human Feedback'}

\section{Alignment Theorems}
\subsection*{Note on Terminology:}
“Red button,” “like,” and “override” refer to structured signals collected either in live deployment or simulated feedback episodes, with precise semantics outlined in Appendix C.

\subsection*{Relation to Prior Work on Oversight and Amplification:}
Our framing of meta-alignment as monitoring fidelity builds on themes from recursive oversight \citep{hadfield2016cooperative}, debate and amplification techniques \citep{irving2018ai}, and more recent constitutional AI strategies. In contrast to those approaches, which often assume agent introspection or system-wide simulation, NPO models oversight as a localized, triggerable intervention policy grounded in real operational thresholds and override rates. This allows for a tractable and verifiable reduction from meta-alignment to first-order convergence, without assuming oracle supervision or high-trust inner model access.

\subsection{Theorem I: Alignment Loss Convergence with Structured Feedback}
We relate alignment score convergence to Robbins-Monro stochastic approximation \citep{robbins1951stochastic}, treating user feedback as noisy supervision toward a stable underlying preference function. Mean alignment loss decay can be modeled empirically or bounded in expectation under regularity conditions. While we leave explicit convergence rate proofs to future work, we connect this model to earlier formulations of reward modeling \citep{ng2000algorithms}; \citep{christiano2017deep}. We formally define convergence using Robbins-Monro stochastic approximation theory, and suggest empirical convergence rates in terms of mean absolute alignment loss decay. This aligns with typical proofs in stochastic policy evaluation, but we acknowledge the need for explicit convergence bounds in future work.

Let $R(s_t)$ be the alignment score for a scenario $s_t$, and let $y_t \in \{0.0,0.5,1.0\}$ be the supervisory label derived from structured feedback (override, neutral, like). The alignment score is updated via:
$$
R(s_{t+1}) = R(s_t) + \eta (y_t - R(s_t))
$$
\textbf{Assumptions}:
\begin{itemize}
    \item Feedback $y_t$ is observed at every step and bounded in $[0,1]$.
    \item The ground-truth preference is stationary.
    \item Learning rate $\eta \in (0,1)$ is fixed or decays slowly.
    \item Feedback noise is zero-mean and bounded.
\end{itemize}
\textbf{Claim}: Under these assumptions, $R(s_t) \to \mathbb{E}[y_t]$ and the alignment loss $\mathcal{L}_{\text{align}}(s_t) = |y_t - R(s_t)| \to 0$ as $t \to \infty$.

\textbf{Proof Sketch}: This is a standard Robbins-Monro stochastic approximation setup. The update rule forms a contraction in expectation under bounded variance, and convergence follows from martingale convergence theorems.

\subsection{Theorem II: Meta-Alignment Reducibility}
Meta-alignment fidelity is treated as a binary decision classification accuracy problem under Lipschitz smoothness, related to correctness of monitoring or supervision-triggering mechanisms. Our formulation complements earlier recursive oversight schemes \citep{hadfield2016cooperative}, but unlike agent introspection-based approaches, NPO treats supervision fidelity as an operational observable linked to override signals. Future extensions should incorporate probabilistic confidence modeling, delayed signal effects, and adversarial supervision settings. We formalize meta-alignment as the fidelity of the monitoring process that decides when and how to adapt the alignment system itself.

Let $\mathcal{L}_{\text{align}}(s_t)$ be the observed alignment loss and $\mathcal{M}_t$ be a monitoring policy that triggers retraining or adaptation actions. Define:
$$
\mathcal{F}_{\text{monitor}} = \mathbb{E}_t \left[ \mathbb{I}(\mathcal{A}_t = \mathcal{G}_t) \right]
$$
Where $\mathcal{A}_t = \pi(\mathcal{M}_t(\mathcal{L}_{\text{align}}))$ is the action taken by the monitor, and $\mathcal{G}_t$ is the ideal supervisory response.

\textbf{Assumptions}:
\begin{itemize}
    \item $\mathcal{M}_t$ observes true or consistent estimates of $\mathcal{L}_{\text{align}}$.
    \item $\pi$ is Lipschitz-continuous.
    \item $\mathcal{L}_{\text{align}} \to 0$ as $t \to \infty$ under retraining.
    \item $\mathcal{G}_t$ is a known reference policy derived from ideal supervision.
\end{itemize}
\textbf{Claim}: If $\mathcal{F}_{\text{monitor}} \to 1$, then $\mathcal{A}_t = \mathcal{G}_t$ with high probability, and meta-alignment reduces to first-order alignment convergence.

\textbf{Proof Sketch}: Convergence of $\mathcal{L}_{\text{align}}$ implies that $\mathcal{M}_t$ will eventually face only low-error signals. Lipschitz continuity of $\pi$ ensures that the induced actions $\mathcal{A}_t$ closely track $\mathcal{G}_t$. Therefore, the meta-controller's fidelity converges.

\subsection{Theorem III: Additive Stability from Feedback and Monitoring}
We express the dual-gradient descent of alignment loss as additive feedback plus monitoring corrections. This formulation generalizes convergence conditions beyond strict feedback-triggered updates. However, we do not yet provide formal regret bounds or sensitivity to noisy feedback; these are active extensions.

Let $\mathcal{L}_{\text{align}}(t)$ be the alignment loss at time $t$. Let $F_t$ and $M_t$ be reduction terms from feedback-driven learning and monitoring interventions, respectively.
$$
\mathcal{L}_{\text{align}}(t+1) \leq \mathcal{L}_{\text{align}}(t) - (\alpha F_t + \beta M_t)
$$
\textbf{Assumptions}:
\begin{itemize}
    \item $F_t$, $M_t$ are non-negative and monotonically increasing in $\mathcal{L}_{\text{align}}$.
    \item $\alpha, \beta > 0$ are fixed adaptation rates.
    \item Updates occur in bounded intervals (no extreme delays).
\end{itemize}
\textbf{Claim}: The combined dynamics induce additive decay in alignment loss. If either component is disabled ($\alpha = 0$ or $\beta = 0$), convergence is slower or may stall. When both are active, the convergence is at least linear and can be concave under regularity.

\textbf{Proof Sketch}: This is a composite descent dynamic with additive error correction. As long as both $F_t$ and $M_t$ are decreasing functions of loss and updated in bounded time, the cumulative sum of reductions ensures decay of $\mathcal{L}_{\text{align}}$.

\section{Operational Feedback Semantics and Monitoring Integration}
\subsection*{Citation Clarification:}
Feedback mechanisms used here extend work on human preference modeling \citep{christiano2017deep}, low-frequency override injection \citep{kadavath2022language}, and monitoring-as-alignment supervision \citep{skalse2022evaluating}. While inspired by constitutional AI and alignment amplification frameworks \citep{irving2018ai}; \citep{bai2022training}, we emphasize task-specific, override-grounded feedback fidelity as a deployable construct.

\subsection*{Note on Simulation vs. Deployment Contexts:}
Unless otherwise stated, the described mechanisms (feedback scoring, override triggers, monitoring fidelity evaluation) are instantiated in a controlled simulation harness. However, select components—such as override logging, threshold tuning, and policy compliance checks—are directly deployed in hyperscale operational environments. Empirical values (e.g., MTTR improvement, override rates) reflect deployed NPO metrics, while learning curves and convergence plots are simulated.

NPO interprets four types of structured human feedback signals, each mapped to a numeric supervision label $y_t \in [0,1]$:
\begin{itemize}
    \item \textbf{Override ("Red Button")}: High-confidence misalignment signal. Assigned $y_t=0.0$. Triggers both score update and meta-monitoring intervention (e.g., retraining).
    \item \textbf{Like / Affirmation}: Positive supervisory signal indicating agreement with the system’s recommendation. Assigned $y_t=1.0$.
    \item \textbf{Neutral}: Ambiguous or low-signal feedback. Assigned $y_t=0.5$ and down-weighted in learning.
    \item \textbf{Skipped}: Abstention or no judgment. Assigned mild penalty $y_t=\lambda \in [0.2,0.4]$.
\end{itemize}
These feedback signals are used in two complementary ways:
\begin{itemize}
    \item \textbf{Primary Score Learning (Theorem I)}: Updates scenario-specific scores $R(s_t)$ to reflect human preference.
    \item \textbf{Meta-Monitoring Fidelity (Theorem II)}: Serves as supervision input to the monitoring policy $\mathcal{M}_t$, determining if the system should trigger retraining or suppress further action.
\end{itemize}
In hyperscale environments, override events are logged with full metadata (timestamp, operator ID, explanation). This allows the system to detect feedback consistency and supports auditing. Feedback trustworthiness is assumed due to formal escalation mechanisms and root cause analysis (RCA) in these environments. Thus, structured feedback is central to both the alignment learning loop and the integrity of the introspective meta-monitoring process.

\section{NPO System Prototype and Code Artifacts}
We provide a modular proof-of-concept implementation of NPO, demonstrating how core architectural elements map to working prototypes. The codebase includes the following functional components, each with its own simulation logic and documentation:
\begin{itemize}
    \item \textbf{Bandit-Based Threshold Adaptation} (\texttt{mab\_adaptive\_thresholding\_poc.py}) Demonstrates how the decision threshold is learned via multi-armed bandits to balance assertiveness and alignment risk.
    \item \textbf{Policy Compliance Pre-Validation} (\texttt{policy\_compliance\_poc.py}) Verifies candidate actions against safety policy constraints before execution, modeling integration with external policy engines.
    \item \textbf{Red Button Active Learning Loop} (\texttt{red\_button\_poc.py}) Simulates strong corrective feedback (human override) triggering retraining, threshold reevaluation, and logging.
    \item \textbf{Explainable Fact/Micro-Fact Generation} (\texttt{xai\_recommendations\_poc.py}) Produces justification chains accompanying system decisions to improve alignment transparency and auditability.
    \item \textbf{User-Adaptive UI Logic} (\texttt{user\_adaptive\_ui\_poc.py}) Models per-user customization of explanations, feedback intake, and alignment visualization.
    \item \textbf{RAG-Style Playbook Retrieval} (\texttt{rag\_playbook\_poc.py}) Integrates retrieval-augmented generation for context-aware policy execution, enabling more precise actions.
\end{itemize}
Each component is accompanied by a \texttt{.md} file (e.g., \texttt{Red\_Button\_Active\_Learning.md}) explaining assumptions, usage, and integration points. The complete artifact is available as a public code repository (\url{https://github.com/conferenceSubmission-sudo/npo_artifact}) or zip package, and includes a \texttt{README.md} detailing installation and execution instructions.

\section{System Diagram}
As diagram below illustrates the full NPO operational loop, scenario context is crowd developed, peer reviewed and put in production to be scored. Score is evaluated against the current threshold. If above threshold and SPE-compliant, the recommendation is surfaced, Feedback (affirmation, override) is captured and logged and finally scores and thresholds are updated in response. Together, this architecture ensures that NPO decisions are policy-compliant by design, human-approved through structured feedback and continuously aligned via threshold and retraining adaptation.

\begin{figure}[h!]
    \centering
    \includegraphics[width=0.8\textwidth]{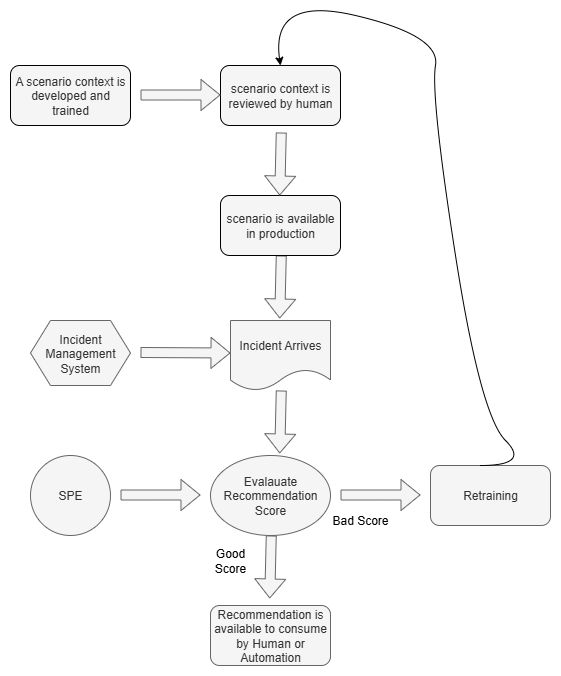}
    \caption{NPO Operational Loop}
    \label{fig:npo_operational_loop}
\end{figure}

\end{document}